\documentclass[runningheads]{llncs}

\usepackage{times}

\usepackage{hyperref}
\usepackage{url}
\makeatletter
\g@addto@macro{\UrlBreaks}{\UrlOrds}
\makeatother

\usepackage{mathtools}
\usepackage{array}
\usepackage{xpatch}
\usepackage{stmaryrd}
\usepackage{bbding}
\usepackage{amssymb}
\usepackage{pifont}
\usepackage{ragged2e}

\newcommand{\cmmnt}[1]{}

\newcommand{\xmark}{\text{\ding{55}}}

\newtheorem{hypothesis}{Hypothesis}

\DeclareMathOperator*{\argmin}{arg\,min}

\newcolumntype{M}[1]{>{\centering\arraybackslash}m{#1}}
\newcolumntype{N}{@{}m{0pt}@{}}

\begin{document}

\title{Using Laplacian Spectrum \\ as Graph Feature Representation}

\author{Edouard Pineau}
\institute{Telecom Paris, Institut Polytechnique de Paris \\ Safran Tech, Signal and Information Technologies}

\maketitle

\begin{abstract}

    Graphs possess exotic features like variable size and absence of natural ordering of the nodes that make them difficult to analyze and compare. To circumvent this problem and learn on graphs, graph feature representation is required. A good graph representation must satisfy the preservation of structural information, with two particular key attributes: consistency under deformation and invariance under isomorphism. While state-of-the-art methods seek such properties with powerful graph neural-networks, we propose to leverage a simple graph feature: the graph Laplacian spectrum (GLS). We first remind and show that GLS satisfies the aforementioned key attributes, using a graph perturbation approach. In particular, we derive bounds for the distance between two GLS that are related to the \textit{divergence to isomorphism}, a standard computationally expensive graph divergence. We finally experiment GLS as graph representation through consistency tests and classification tasks, and show that it is a strong graph feature representation baseline.
\end{abstract}

\section{Introduction}
\label{sec:introduction}

No matter where and at which scale we look, graphs are present. Social networks, public transport, information networks, molecules, any structural dependency between elements of a global system is a graph. An important task is to extract information from these graphs in order to understand whether they contain certain structural properties that can be represented and used in downstream machine learning tasks. In general, graphs are difficult to use as input of standard algorithms because of their exotic features like variable size and absence of natural orientation. Consequently, graph feature representation with equal dimensionality and dimension-wise alignment is required to learn on graphs. 

We know that any embedding method must satisfy the preservation of structural information, and in particular for graph must satisfy two key attributes: consistency under deformation and invariance under isomorphism. The first forces the embedding to discriminate two graphs consistently with their structural dissimilarity. The second enables to have one representation for one graph, which can be a challenge since one graph has many possible orientations. In this paper, we propose to analyze the importance of satisfying the introduced criteria through a known, simple, expressive and efficient candidate graph feature representation: the graph Laplacian spectrum (GLS). 

The Laplacian matrix of a graph is a major object in spectral learning \cite{belkin2002laplacian}. However, most of the attention is usually directed to its eigenvectors and not its spectrum, and spectral learning is generally applied to node clustering or classification, not whole-graph representation. But, GLS holds interesting properties for graph representation. First, the Laplacian eigenvalues give many structural information like the presence of communities and partitions \cite{newman2013spectral}, the regularity, the closed-walks enumeration, the diameter or the connectedness  of the graph \cite{brouwer2011spectra}. It is also interpretable in term of signal processing \cite{shuman2016vertex} or mechanics \cite{bonald2018weighted}. Second, it is backed by efficient and robust approximate eigen-decomposition algorithms enabling to scale on large graphs and huge datasets \cite{halko2011finding}. Third, GLS is invariant under graph isomorphism. Finally, each eigenvalue of the GLS can be seen as an graph feature representation by itself, containing specific structural information. Hence any subset of Laplacian eigenvalues is a meaningful and valuable embedding. This enables the usage of truncated Laplacian spectrum (t-GLS) instead of GLS as whole-graph feature representation. Using t-GLS reduces the embedding time thanks to eigenvalue algorithms that do not require entire diagonalization to give partial spectrum \cite{halko2011finding}.

These properties tell us that GLS is a good graph feature representation candidate. In this paper we go further and analyze the interesting properties of the Laplacian spectrum through the following contributions: (1) we build a perturbation-based framework to analyze the representation capacity of the GLS, (2) we analyze the consistency between structural deformation of the graph and its GLS by deriving bounds for the distance between the GLS of two graphs, (3) we validate the consistency and the representational power of the GLS with different experiments on synthetic and real graphs.

The rest of the paper is built as follows. A presentation of the mathematical framework and the theoretical analysis are displayed respectively in Section \ref{sec:problemsetup} and \ref{sec:perturbation}. Section \ref{sec:experiments} proposes experiments to illustrate theoretical results and show the representational power of GLS. Finally, Section \ref{sec:related_work} describes related work about graph representation.

\section{Perturbation approach and problem setup}
\label{sec:problemsetup}

We consider two undirected and weighted graphs $G_1=(V_1, E_1, W_1)$ and $G_2=(V_2, E_2, W_2)$ with respective adjacency matrix $W_1$ and $W_2$, degree matrix $D_1$ and $D_2$. These matrices are set with respect to an arbitrary indexing of the nodes. Laplacian matrix $L_i$ of $G_i$ is defined as $L_i = D_i-W_i$. We aim at using the GLS to build fixed-dimensional representation that encodes structural information to compare any graphs $G_1$ and $G_2$ that are not aligned nor equally sized. For the rest of the paper, and without loss of generality we postulate that $|V_1| \leq |V_2|$. The rest of this section introduces the definitions, hypothesis and notations needed for our theoretical analysis of the GLS.

\begin{definition}
\label{def:perturbation}
Let $G=(V, E, W)$ a weighted graph with $n$ nodes, with $W \in \mathcal{M}_{n \times n}$ the $n \times n$ weighted adjacency matrices. We define $P \in \mathcal{M}_{n \times n}$ a symmetric matrix with $P_{ii}=0$, $P_{ij} \in \left[ -W_{ij}, W_{ij} \right]$ such that $W_{ij}+P_{ij} \in \left[ 0, 1 \right] \forall (i,j)$. We define the two following perturbations applied on graph $G$:

\begin{itemize}
\item Adding isolated nodes: $\overline{W} = \begin{bmatrix}
               W & 0_{n \times m} \\
               0_{m \times n} & 0_{m \times m}
    \end{bmatrix}$

\item Adding or removing edges: ${W}^P = W + P$
\end{itemize}

We call \textit{edge-perturbation} the addition or removal of edges, and \textit{node-perturbation} the addition of nodes. A complete perturbation is done by adding isolated nodes and perturbing the augmented graph with edge-perturbation. We note that the withdrawal of a node is equivalent to removal of all edges around this nodes. Moreover, if graph $G$ is unweighted, i.e. with binary adjacency, then edge perturbations $P_{ij}\in \{-1, 0, 1\}$.

\end{definition}

\begin{remark}

If $P=-W+\Pi^T W \Pi$ with $\Pi \in \mathcal{P}(n)$ then the perturbation is a permutation of the node indexing. Such a perturbation is not interesting and edge perturbation due to node indexing has to be annihilated by a permutation matrix as in the following definition.

\end{remark}

\begin{definition}
\label{def:sparseperturbation}

    We say that $G^{P^*}$ is a perturbed version of $G$ if we have

    \begin{equation*}
        \begin{cases}
            W^{P^*} = {\Pi^*}^T (\overline{W} + P^*) \Pi^* \\
            s.t. \ \Pi^* = \argmin_{\Pi} { \lVert W^{P^*} - {\Pi}^T \overline{W} \Pi \rVert _1}
        \end{cases}
    \end{equation*}

    i.e. such that $P^*$ is the sparsest possible i.e. does not include permutations.
\end{definition}

\paragraph{Notations} We denote $P^*$ \textit{the sparsest perturbation} as defined in Definition \ref{def:sparseperturbation}. We denote $\overline{G}$ the completion of G with isolated nodes. If $M$ is a matrix associated to $G$, we denote $\overline{M}$ the equivalent matrix for $\overline{G}$. We denote $\lambda(X)$ the eigenvalue of a square matrix $X$ in ascending order, $\lambda_i(X)$ the $i^{th}$ smallest eigenvalue.

\begin{hypothesis}
\label{hyp:perturbed_subgraph}
Without loss of generality, we assume that $G_2$ is a perturbed version of $G_1$, i.e. $\exists P^*$ the sparsest $|V_2|$-square perturbation matrix and $\Pi^* \in \mathcal{P}(|V_2|)$ a $|V_2|$-square permutation matrix such that $W_2 = {\Pi^*}^T \left( \overline{W_1} + P^* \right) \Pi^*$. $P^*$ is a $|V_2|$-square block matrix, with top-left block $P^*_{11}$ being a $|V_1|$-square perturbation matrix for graph $G_1$. Bottom right block $P^*_{22}$ is the $(|V_2|-|V_1|)$-square adjacency matrix of the additional nodes. $P^*_{12}$ is the $|V_1| \times (|V_2|-|V_1|)$ adjacency matrix representing the links between graph $G_1$ and the additional nodes $V_2 \setminus V_1$. 

\end{hypothesis}

We have defined a notion of continuous deformation of graphs. This deformation has a natural and simple interpretation: any graph $G_2$ is a perturbed version of graph $G_1$, and the larger the perturbation the higher the structural dissimilarity between $G_1$ and $G_2$. 

The next section uses the previously presented mathematical framework to analyze the consistency of the Laplacian spectrum as graph representation and its natural link to graph isomorphism problem. 

\section{Laplacian spectrum as graph feature representation}
\label{sec:perturbation}

We place ourselves under the Hypothesis \ref{hyp:perturbed_subgraph} saying that the difference between graphs $G_1$ and $G_2$ is characterized by the unknown deformation $P^*$. A good embedding of these graphs should be close when level of deformation is low, and far otherwise. This level of deformation can be quantified by the global and node-wise entries of $P^*$. These features are by construction present in the Laplacian of $P^*$, denoted $L_{P^*}$. We use this idea to propose an analysis of the distance between to GLS. 

All proofs are detailed in appendix.

\subsection{Consistency under deformation and relation to graph isomorphism}

Two graphs $G_1$ and $G_2$ are isomorphic if and only if $\exists \Pi \in \mathcal{P}(|V_1|)$ such that ${L}_2 = \Pi^{-1}{L}_1\Pi$ \cite{merris1994laplacian}, hence when they are structurally equivalent irrespective to the vertex ordering. Several papers has proposed to use a notion of \textit{divergence to graph isomophism} (DGI) to compare graphs \cite{grohe2018graph,rameshkumar2013laplacian}. The DGI between graphs $G_1$ and $G_2$ is generally the minimal Frobenius norm of the difference between ${L}_1$ and $\Pi^{-1}{L}_2\Pi$. Considering this definition, the following Lemma links the graph-isomorphism problem and the Laplacian of the hypothetical perturbation $P^*$ and show that this divergence is the norm of $L_{P^*}$:

\begin{lemma}
\label{lemma:perturbedlaplacian}

Using the notations from Hyp. \ref{hyp:perturbed_subgraph}, we have $L_2 = {\Pi^*}^T \left(\overline{L_1} + L_{P^*}\right) {\Pi^*}$, with $L_{P^*} = \text{diag}(P^* \mathbf{1}_{|V_1|}) -  {P^*}$ the Laplacian of $P^*$ and $\mathbf{1}_n$ the $n$-dimensional unit vector. In particular, $\min_\Pi \lVert {L_2 - {\Pi}^T \overline{L_1} {\Pi}}\rVert _F = \lVert L_{P^*} \rVert _F$. 

\end{lemma}

We remind that graph isomorphism is at best solved in quasipolynomial time \cite{babai2016graph} and can not be used in practice for large graphs and datasets. The following Proposition show how the distance between GLS relaxes the isomorphism-based graph divergence. 

\begin{proposition} 
\label{proposition:weyl}
Using Hypothesis \ref{hyp:perturbed_subgraph} and Lemma \ref{lemma:perturbedlaplacian}: $\lVert \lambda(L_2) - \lambda(\overline{L_1}) \rVert_2 \leq \lVert L_{P^*} \rVert _F$.

\cmmnt{
    \begin{align*}
        \lVert \lambda(L_2) - \lambda(\overline{L_1}) \rVert_2 \leq \lVert L_{P^*} \rVert _F. 
    \end{align*}
}

\end{proposition}

The above result tells us that the higher the difference between GLS, the larger the hypothetical perturbation $P^*$, i.e. the higher the structural dissimilarity. 

We now study the implication of GLS closeness. This problem tackles the notion of \textit{non-isomorphic $L$-cospectrality}, i.e. the idea that two graphs can have equal eigenvalues while having different Laplacian matrix \cite{brouwer2011spectra}. The following proposition gives a simple insight into the problem of spectral characterization in our perturbation-based framework:

\begin{proposition}
\label{prop:upper_bound_perturbation}
We denote $L_i=Q_i \Lambda_i Q_i^T$ the singular value decomposition (SVD) of $L_i$ such that the diagonal of $\Lambda_i$ is in ascending order. Therefore we have the inequality $\lVert L_{P^*} \rVert_F \leq \lVert \lambda(L_2) - \lambda(\overline{L_1}) \rVert_2 + \lVert {\Pi^*}^T \overline{Q_1} \Lambda_2 \overline{Q_1}^T {\Pi^*} - L_2 \rVert_F$.

\cmmnt{
\begin{align*}
    \lVert L_{P^*} \rVert_F
        &\leq \lVert \lambda(L_2) - \lambda(\overline{L_1}) \rVert_2 + \lVert {\Pi^*}^T \overline{Q_1} \Lambda_2 \overline{Q_1}^T {\Pi^*} - L_2 \rVert_F.
\end{align*}
}

\end{proposition}

This proposition shows that equal spectrum means equal graphs only when eigenvectors are also equal. Otherwise, $L$-cospectrality for non-isomorphic graphs tells us that there exists families of graphs that are not fully determined by their spectrum. These families are characterized by some structural properties such that two non-isomorphic graphs with equal Laplacian spectrum share these properties but not their adjacency \cite{van2003graphs}. In practice, this is not a problem. First, almost all graphs are determined by their spectrum \cite{brouwer2011spectra}. Second, equal GLS indicates the precious information that graphs share common structural properties, no matter the adjacency matrix. These properties might be what we seek to represent when representing graphs for ML tasks. Third, non-isomorphic $L$-cospectrality concerns equally sized graphs which is not likely with respect to all possible real-life graphs. When the studied dataset contains specifically $L$-cospectral non-isomorphic graphs and when the task requires unique representation property, GLS is not appropriate and more sophisticated and powerful embedding methods taking for example eigenvectors into account \cite{verma2017hunt} should be studied and used. Otherwise, i.e. in almost every situations, according to previously presented results, GLS characterizes the graph and is directly related to the hypothetical perturbation $P^*$. 

Nevertheless, we accordingly propose the Proposition \ref{prop:upper_bound} to better understand GLS proximity even when graphs are non-isomorphic cospectral.

\begin{proposition}
\label{prop:upper_bound}
The closer the GLS, the closer to unitary-similarity the Laplacian matrices.

\end{proposition}

We remind that two real $n$-square matrices $A$ and $B$ are \textit{unitary-similarity} if there exists an orthogonal matrix $O$ such that $B=OAO^{T}$. Similarity is an \textit{equivalence relation} on the space of square-matrices. Moreover, divergence to unitary-similarity is a relaxed version of the divergence to graph-isomorphism \cite{grohe2018graph}, where the permutation matrix space is replaced by a unitary matrix space. Finally from Proposition \ref{proposition:weyl} and \ref{prop:upper_bound} we can bound the distance between GLS as follows: 

\begin{align*}
    \min_{O \in \mathcal{O}(|V_2|)} \lVert \overline{L_1} - O L_2 O^{T} \rVert_F \leq {\|{\lambda(\overline{L_1}) - \lambda(L_2)}\|_2} \leq \lVert L_{P^*} \rVert
\end{align*}

In this section, we have shown that structural similarity (divergence) between graphs can be reasonably approximated by the similarity (divergence) between their GLS. 

\subsection{Laplacian spectrum as whole-graph representation in practice}
\label{subsec:completion}

Previous section showed the capacity of the distance between Laplacian spectrum to serve as proxy for graph similarity. In practice, a fixed embedding dimension $d$ must be chosen for all graphs in dataset $\mathcal{D}$. According to previous analysis, the most obvious dimension is $d=\max_{G \in \mathcal{D}}{|V|}$ and all graphs with less than $d$ nodes may be padded with isolated nodes. We note that padding with isolated nodes is equivalent than adding zeros in the GLS. Nevertheless, in some datasets, some graphs can be significantly larger and the padding can become abusive. We therefore propose for these graph to have $d<\max_{G \in \mathcal{D}}{|V|}$. We simply truncate the GLS such that we keep only the highest $d$ eigenvalues. This method also enables to save computation time.

The problem with this method is that we may lose information for graphs with more than $d$ nodes. In practice, for large graphs, the contribution of the lowest eigenvalues to the distance between GLS as a proxy for graph divergence is negligible. In particular, large graph have many sparse areas, such that many eigenvalues are very low, hence truncating the bottom part of the GLS may not be a problem. We assess the impact of the truncation in the experimental section.  

Though, we can also propose several ways to avoid this problem, like embedding the lowest eigenvalues with simple statistics, like moments or histograms. In the experimental section, we do not use this trick.

\section{Experiments}
\label{sec:experiments}

All experiments can be reproduced using the code provided at the following address: \url{https://github.com/edouardpineau/Using-Laplacian-Spectrum-as-Graph-Feature-Representation}

\subsection{Preliminary experiments} 
\label{subsec:preliminary_experiments}
As a first illustration of deformation-based results presented in Section \ref{sec:perturbation}, we propose to use Erdos-R\'{e}nyi random graphs \cite{erdds1959random} with parameter $p=0.05$. We focus on three simple experiments. 

\begin{figure}[h]
    \centering
    \includegraphics[width=0.49\textwidth]{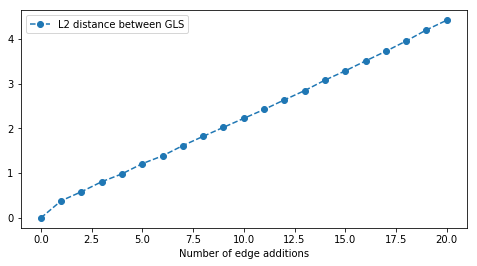}
    \includegraphics[width=0.49\textwidth]{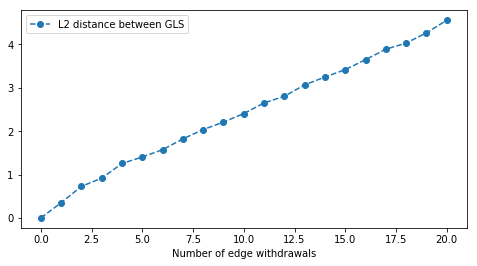}
    \caption{Experimental results to illustrate how GLS behaves under edge addition (left), edge withdrawal (right). In this case, studied adjacency and perturbation matrix are binary.}
    \label{fig:edge_perturbations}
\end{figure}

First, the distance between the Laplacian spectrum of a graph and a perturbed version of this graph is related to the number of perturbations. We can find the experimental illustration in Figure \ref{fig:edge_perturbations} (similar to those in \cite{wilson2008study}). We see that the number of perturbations is directly related to the distance between GLS features for edge addition and edge withdrawal. A relation between graph sparsity and Laplacian eigenvalues can be seen for example through the Gershgorin circle theorem \cite{gershgorin1931uber}.

Second, we mentioned that when a graph is significantly bigger than other graphs of a dataset, we can use a truncated GLS (t-GLS). This method both saves computation time thanks to iterative eigenvalues algorithms and avoids the addition of isolated nodes in all other graphs. In Figure \ref{fig:node_perturbations}, we show results of experiments showing that t-GLS is consistent with node addition. As experimental setup, we take a reference graph with $n$ nodes and compute its GLS. Then we add a randomly connected node and compute the t-GLS of the new graph, by keeping only the $n$ largest eigenvalues. We repeat it 20 times. We compute the $L_2$-distance to reference GLS, for different levels of connectivity for the additional nodes. We first observe that the t-GLS is consistent with node addition. We also confirm our previous theoretical results by observing that the more connected the additional nodes, the higher the GLS divergence.

\begin{figure}[h]
    \centering
    \includegraphics[width=0.49\textwidth]{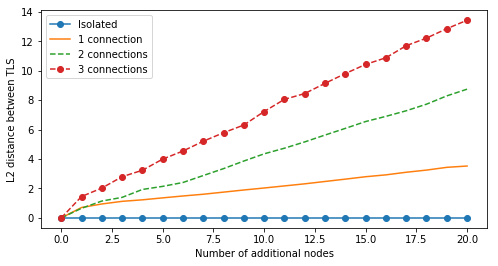}
    \includegraphics[width=0.49\textwidth]{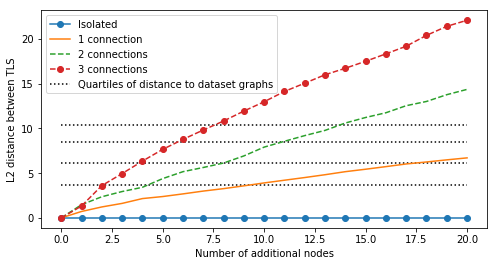}
    \caption{Experimental results illustrate how truncated GLS behaves under iterative addition of 20 new nodes with respectively 0, 1, 2 and 3 random connections with graph, for respectively synthetic a 80-nodes Erdos-Reyni graph (left) and a 28-nodes molecular graph from MUTAG dataset (right). Horizontal dotted lines (right figure) are the quartiles 25, 50, 75 and 100 of the distances between the GLS of the 28-nodes graph and the other 187 graphs of the dataset.}
    \label{fig:node_perturbations}
\end{figure}

\subsection{Classification of molecular and social network graphs}

We evaluate spectral feature embedding with a classification task on molecular graphs and social network graphs. Experimental setup for classification task is given in Appendix \ref{app:experimental_setup}. We assume here that two structurally close graphs belong to the same class. We challenge this assumption with the following experiments.

We propose to compare GLS-based classification results to those obtained by feature-based and deep learning methods. Standard graph feature representation methods are: Earth Mover's Distance \cite{nikolentzos2017matching} (EMD), Pyramid Match \cite{nikolentzos2017matching} (PM), Feature-Based \cite{barnett2016feature} (FB) and Dynamic-Based Features \cite{gomez2017dynamics} (DyF). All of these methods use support vector classifier (SVC) over extracted features. Deep learning methods are: Variational Recurrent Graph Classifier \cite{pineau2019variational} (VRGC), Graph Convolutional Network \cite{kipf2016semi} (GCN), Deep Graph CNN \cite{zhang2018end} (DGCNN), Capsule GNN \cite{xinyi2018capsule} (CapsGNN), Graph Isomorphism Network \cite{xu2018powerful} (GIN) and GraphSAGE \cite{hamilton2017inductive}. All deep learning methods are end-to-end graph classifers. A description of these models is given in the related work, Section \ref{sec:related_work}.

All values reported in Table \ref{tab:results_molecular} and Table \ref{tab:results_social_deep} are taken from the above-mentioned papers. 

\paragraph{Molecular graphs}
\label{subsec:molecular_classification}

We use five datasets for the experiments: Mutag (MT), Enzymes (EZ), Proteins Full (PF), Dobson and Doig (DD) and National Cancer Institute (NCI1) \cite{KKMMN2016}. All graphs are chemical components. Nodes are atoms or molecules and edges represent checmical or electrostatic bindings. We note that molecular graphs contain node attributes, that are used by some models presented in Table \ref{tab:results_molecular}. We let the question of the relevance of comparing models with slightly different inputs to the discretion of the reader. Description and statistics of molecular datasets are presented in Table \ref{tab:molecular_datasets}, Appendix \ref{app:datasets_characteristics}.

\begin{table}[h]
  \begin{center}
  \renewcommand{\arraystretch}{1}
    \begin{tabular}{l|M{1.8cm} M{1.8cm} M{1.8cm} M{1.8cm} M{1.8cm} M{1.8cm}}
                    & MT   &   EZ       & PF      & DD      & NCI1 \\
        \hline
        EMD + SVC           & 86.1 $\pm$ 0.8 &      36.8 $\pm$ 0.8   & -     &   -   & 72.7 $\pm$ 0.2 \\
        PM + SVC            & 85.6 $\pm$ 0.6 &      28.2 $\pm$ 0.4   & -     &  75.6 $\pm$ 0.6 & 69.7 $\pm$ 0.1 \\
        FB + SVC            & 84.7 $\pm$ 2.0 &      29.0 $\pm$ 1.2   & 70.0 $\pm$ 1.3  &   -   & 62.9 $\pm$ 1.0 \\
        DyF + SVC           & 86.3 $\pm$ 1.3 &      26.6 $\pm$ 1.2   & 73.1 $\pm$ 0.4 &   -   & 66.6 $\pm$ 0.3 \\
        FGSD + SVC        & 92.1 &    -     & 73.4  & 77.1  & 79.8 \\
        \hline
        VRGC          & 86.3 $\pm$ 8.6 &      48.4 $\pm$ 6.2   & 74.8 $\pm$ 3.0   &   -  & 80.7 $\pm$ 2.2 \\
        GCN*          & 85.6 $\pm$ 5.8 &      -                & 76.0 $\pm$ 3.2    &   -  & 80.2 $\pm$ 2.0 \\
        DGCNN*        & 85.8 $\pm$ 1.7 &          51.0 $\pm$ 7.3   & 75.5 $\pm$ 0.9 & 79.4 $\pm$ 0.9 & 74.4 $\pm$ 0.5 \\
        CapsGNN*      & 86.7 $\pm$ 6.9 &      54.7 $\pm$ 5.7   & 76.3 $\pm$ 3.6 & 75.4 $\pm$ 4.2 & 78.4 $\pm$ 1.6 \\
        GIN-0*        & 89.4 $\pm$ 5.6 &      -                & 76.2 $\pm$ 2.8    &   -  & 82.7 $\pm$ 1.7 \\
        GraphSAGE*    & 85.1 $\pm$ 7.6 &      -                & 75.9 $\pm$ 3.2    &   -  & 77.7 $\pm$ 1.5 \\
        \hline
        GLS + SVC         & 87.9 $\pm$ 7.0 & 40.7 $\pm$ 6.3 & 75.3 $\pm$ 3.5 &  74.3 $\pm$ 3.5 & 73.3 $\pm$ 2.1 \\
      \end{tabular}
      \renewcommand{\arraystretch}{4}
  \end{center}
  \caption{Accuracy ($\%$) of classification with different graph representations, on molecular graphs. SVC stands for support vector classifier. Comparative models are divided into two groups: feature + SVC and end-to-end deep learning. *Models using node attributes.}
  \label{tab:results_molecular}
\end{table}

\paragraph{Social network graphs}
\label{subsec:social_classification}

We use five datasets for the experiments: IMDB-Binary (IMBD-B), IMDB-Multi (IMDB-M), REDDIT-Binary (REDDIT-B), REDDIT-5K-Multi (REDDIT-M) and COLLAB. All graphs are social networks. The graphs of these datasets do not contain node attributes. Therefore, we can more appropriately compare GLS + SVC to deep learning based classification. Statistics about social networks datasets are presented in Table \ref{tab:social_datasets}, Appendix \ref{app:datasets_characteristics}.

\begin{table}[h]
\selectfont
  \begin{center}
  \renewcommand{\arraystretch}{1}
    \begin{tabular}{l|M{1.7cm} M{1.7cm} M{1.7cm} M{1.7cm} M{1.7cm}}
                    & IMDB-B    & IMDB-M   & REDDIT-B    & REDDIT-M    & COLLAB \\
      \hline
      GCN                   & 74.0 $\pm$ 3.4  & 51.9 $\pm$ 3.8  & -  & -  & 79.0 $\pm$ 1.8   \\
      DGCNN                 & 70.0 $\pm$ 0.9   &  47.8 $\pm$ 0.9   &  76.0 $\pm$ 1.7   &  - & 73.8 $\pm$ 0.5\\
      CapsGNN               & 73.1 $\pm$ 4.8   &  50.3 $\pm$ 2.7 & - &  52.9 $\pm$ 1.5  & 79.6 $\pm$ 0.9 \\
      GIN-0                 & 75.1 $\pm$ 5.1 & 52.3 $\pm$ 2.8 & 92.4 $\pm$ 2.5 &  57.5 $\pm$ 1.5 & 80.2 $\pm$ 1.9 \\
      GraphSAGE             & 72.3 $\pm$ 5.3 & 50.9 $\pm$ 2.2    & -     & -    &   -   \\
      \hline
      GLS + SVC             & 73.2 $\pm$ 4.2 & 48.5 $\pm$ 2.5 & 87.4 $\pm$ 3.4 & 52.0 $\pm$ 1.8 &  78.5 $\pm$ 1.1 \\
    \end{tabular}
      \renewcommand{\arraystretch}{4}
  \end{center}
  \caption{Classification accuracy ($\%$) of different deep learning based models plus ours over standard social networks datasets. Graphs of these datasets does not have node features. SVC stands for support vector classifier.}
  \label{tab:results_social_deep}
\end{table}

\paragraph{Analysis of the results}

The classification results above illustrate the capacity of GLS to capture graph structural information, under the assumption that structurally close graphs belong to the same class. The graph neural-networks are globally more expressive since they can leverage specific information for graph classification since is end-to-end. In particular, they obtain strong results when there are node labels (see molecular experiments \ref{subsec:molecular_classification}). Nevertheless, GLS is a simple way to represent graphs in an unsupervised manner, with theoretical background, simplicity of implementation (eigendecomposition is accessible to anyone interested in any computer) and competitive downstream classification results. 

\paragraph{On the reasonability of using truncated GLS}

We assess the impact of truncating the GLS. Using truncated GLS (t-GLS) enables to (1) reduce the computational cost for large graphs and (2) reduce the dimensionality of the graph representation for all graphs. Results are presented in Figure \ref{fig:node_perturbations} for molecular datasets. 

\begin{figure}[h]
    \centering
    \includegraphics[width=0.7\textwidth]{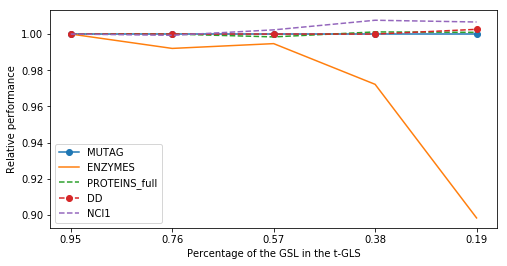}
    \caption{Illustration of the impact of the truncation in term of classification accuracy of the molecular graphs. We represent the impact relatively to the 95-percentile truncation adopted for classification experiments.}
    \label{fig:node_perturbations}
\end{figure}

We see that truncating GLS is not highly impacting classification results. Only ENZYMES multi-class classification, which is a particularly difficult task (see experiments in Section \ref{subsec:molecular_classification}), suffers from truncation. Additional insight about the t-GLS is given in Appendix \ref{app:truncation}.

\paragraph{Computation analysis}

GLS extraction is a quick task, thanks to very efficient eigendecomposition algorithms for sparse graph matrices \cite{halko2011finding}. For example the complete set of molecular experiments (embedding + SVM) took approximately 5 minutes on a single CPU, most of it dedicated to the computation of largest graphs of DD.

\section{Related work}
\label{sec:related_work}

We propose to divide graph feature representation into three categories: graph kernel methods, feature-based methods and deep learning. 

\paragraph{Graph kernel methods}
Kernel methods create a high-dimensional feature representation of data. The \textit{kernel trick} \cite{shawe2004kernel} avoids to compute explicitly the coordinates in the feature space, only the inner product between all pairs of data image: it is an \textit{implicit embedding methods}. These methods are applied to graphs \cite{nikolentzos2018degeneracy,nikolentzos2017matching}. It consists in performing pairwise comparisons between atomic substructures of the graphs until a good representative dictionary is found. The embedding of a graph is then the number of occurrences of these substructures within it. These substructures can be graphlets \cite{yanardag2015deep}, subtree patterns \cite{shervashidze2011weisfeiler}, random walks \cite{vishwanathan2010graph} or paths \cite{borgwardt2005shortest}. The main difficulty lives in the choice of appropriate algorithm and kernel that accept graphs with variable size and capture useful feature for downstream task. Moreover, kernel methods can be computationally expensive but techniques like the Nyström algorithm \cite{williams2001using} allow to lower the number of comparison with a low rank approximation of the similarity matrix.

\paragraph{Feature-based methods}
Feature-based representation methods \cite{barnett2016feature} represent each graph as the concatenation of features. Generally, the feature-based representation can offer a certain degree of interpretability and transparency. The most basic ones are the number of nodes or edges, the histogram of node degrees. These simple graph-level features offers by construction the sought isormorphism-invariance but suffer from low expressiveness. More sophisticated algorithms consider features based on attributes of random walks on the graph \cite{gomez2017dynamics} while others are graphlet based \cite{kondor2009graphlet}. \cite{kondor2008skew} explicitly built permutation-invariant features by mapping the adjacency matrix to a function on the symmetric group. \cite{verma2017hunt} proposed a family of graph spectral distances to build graph features. Experimental work in \cite{de2018simple} used normalized Laplacian spectrum with random forest for graph classification with promising results. \cite{wilson2008study} analyzes the cospectrality of different graph matrices and studies experimentally the representational power of their spectra. These two last works are directly related to the current work. Nevertheless, in both cases, the theoretical analysis is absent and comparative experiment with current benchmarks and methods is limited. in this paper we propose a response to these concerns. 

\paragraph{Deep learning based methods}
GNNs learn representation of nodes of a graph by leveraging together their attributes, information on neighboring nodes and the attributes of the connecting edges. When graphs have no vertex features, the node degrees are used instead. To create graph-level representation instead of node representation, node embeddings are pooled by a permutation invariant readout function like summation or more sophisticated information preserving ones \cite{ying2018hierarchical,zhang2018end}. A condition of optimality for readout function is presented in \cite{xu2018powerful}. Recently, \cite{xinyi2018capsule} levraged capsule networks \cite{sabour2017dynamic}, neural units designed to enable to better preserve information at pooling time. Other popular evolution of GNNs formulate convolution-like operations on graphs. Formulation in spectral domain \cite{bruna2013spectral,defferrard2016convolutional} is limited to the processing of different signals on a single graph structure, because they rely on the fixed spectrum of the Laplacian. Conversly, formulation in spatial domain are not limited to one graph structure \cite{atwood2016diffusion,duvenaud2015convolutional,niepert2016learning,hamilton2017inductive} and can infer information from unseen graph structures. At the same time, alternative to GNN exist and are related to random walk embedding. In \cite{li2017deepcas}, neural networks help to sample paths which preserve significant graph properties. Other approaches transforms graphs into sequence of nodes embedding passed into a recurrent neural network (RNN) \cite{you2018graphrnn,pineau2019variational} to get useful embedding. These models do not inherently include isomorphism-invariance but greedy learn it by seeing the same graph numerous times with different node ordering and embedding. \cmmnt{We finally note that recent work mixed kernel method and neural networks with application to graph embedding \cite{laforgue2018autoencoding}. }These methods are powerful and globally obtain a high level of expressiveness (see experimental section \ref{subsec:molecular_classification}).

\section{Conclusion}
\label{sec:conclusion}

In this paper, we analyzed the graph Laplacian spectrum (GLS) as whole graph representation. In particular, we showed that comparing two GLS is a good proxy for the divergence between two graphs in term of structural information. We coupled these results to the natural invariance to isomorphism, the simplicity of implementation, the computational efficiency offered by modern randomized algorithms and the rare occurrence of detrimental $L$-cospectral non-isomorphic graphs to propose the GLS as a strong baseline graph feature representation.


\bibliographystyle{splncs04}
\bibliography{mybibliography}

\clearpage
\appendix

\section{Proof of Lemma \ref{lemma:perturbedlaplacian}}
\label{proof:lemma:perturbedlaplacian}

\begin{align*}
    L_2
    &= D_2 - W_2  \\
    &= \text{diag}( W_2 \mathbf{1} ) - W_2 \\
    &= \text{diag}( {\Pi^*}^T \left( \overline{W_1} + P^* \right) \Pi^* \mathbf{1} ) - {\Pi^*} \left( \overline{W_1} + P^* \right) {\Pi^*} \\
    &= \text{diag}( {\Pi^*}^T \overline{W_1} \Pi^* \mathbf{1} ) + \text{diag}( {\Pi^*}^T P^* \Pi^* \mathbf{1} ) - {\Pi^*}^T \overline{W_1} {\Pi^*} - {\Pi^*}^T  P^* {\Pi^*} \\
    &= {\Pi^*}^T \overline{D_1} \Pi^* - {\Pi^*}^T \overline{W_1} \Pi^* + {\Pi^*}^T D_{P^*} \Pi^* - {\Pi^*}^T P^* \Pi^* \\
    &= {\Pi^*}^T \overline{L_1} {\Pi^*} + {\Pi^*}^T L_{P^*} {\Pi^*}
\end{align*}

with $L_{P^*} = \text{diag}(P^*\mathbf{1}) - {P^*} = D_{P^*} -  {P^*}$ and $\mathbf{1}$ the unit vector.

Therefore, 

\begin{align*}
    \min_\Pi \lVert {L_2 - {\Pi}^T \overline{L_1} {\Pi}}\rVert _F
    = \min_\Pi \lVert {{\Pi}^T L_{P^*} {\Pi}}\rVert _F
    = \lVert L_{P^*} \rVert _F
\end{align*}

\cmmnt{

\section{Proof of invariance to isomorphism}
\label{proof:lemma:perminvariant}
    
Let $G=(V, E, W)$ be an undirected and weighted graph, $L$ its Laplacian matrix and $\Pi\in \mathcal{P}(|V|)$ be a permutation matrix. A permutation of node indexing implies a permutation of both rows and columns of the Laplacian matrix. 

The spectrum of $L$ is the set of roots of $P(\lambda)=det\left(L-\lambda I_{|V|}\right)$. We want the spectrum of $\Pi^+ L \Pi$:

\begin{align*}
    P^{\Pi}(\lambda)
    &= det\left(\Pi^T L \Pi-\lambda I_{|V|}\right) \\
    &= det\left(\Pi^T L \Pi-\lambda \Pi^+ \Pi \right) \\
    &= det\left(\Pi^T \left(L -\lambda I_{|V|} \right)\Pi \right) \\
    &= det\left(\Pi^T \right) det\left(L -\lambda I_{|V|} \right) det\left( \Pi \right) \\
\end{align*}

Yet we know that the determinant of a matrix is invariant to transpose and that the determinant of a permutation matrix is equal to its signature: \\

\begin{align*}
    det\left( \Pi \right)=det\left( \Pi^T \right) = (-1)^{|V| - \sum_{i=1}^{|V|}{\Pi_{ii}} }
\end{align*}
 
 Hence:
 
\begin{align*}
    P^{\Pi}(\lambda)
    &= (-1)^{2 \left(|V| - \sum_{i=1}^{|V|}{\Pi_{ii}} \right)} det\left(L -\lambda I_{|V|} \right) \\
    &= det\left(L -\lambda I_{|V|} \right) \\
    &= P(\lambda)
\end{align*}
}

\section{Proof of Proposition \ref{proposition:weyl}}
\label{proof:proposition:weyl}

From lemma \ref{lemma:perturbedlaplacian} we have $ L_2 = {\Pi^*}^T \overline{L_1} {\Pi^*} + {\Pi^*}^T L_{P^*} {\Pi^*}$. Moreover, from Weyl's eigenvalues inequalities and since eigenvalues are isomorphism invariant:

\begin{equation*}
    \begin{cases*}
        \lambda_{i}(L_2) \leq \lambda_{i}(\overline{L_1}) + \lambda_{|V_2|}(L_{P^*}) \\ 
        \lambda_{i}(\overline{L_1}) + \lambda_1(L_{P^*}) \leq \lambda_{i}(L_2)
    \end{cases*}    
\end{equation*}

\noindent Hence: $\lambda_1(L_{P^*}) \leq \lambda_{i}(L_2) - \lambda_{i}(\overline{L_1}) \leq \lambda_{|V|}(L_{P^*})$.

Now let $(\lambda, x)$ be any eigen couple of a matrix $M\in \mathcal{M}_{n\times n}$. We can always pick $i \in \{1\dots n\}$ and build $x$ such that $|x_i|=1$ and $|x_{j\neq i}|<1$. Hence:

\begin{align*}
    (Mx)_i = \lambda x_i
    &\Longleftrightarrow \sum_{j=1}^n{m_{ij}x_j} = \lambda x_i \\
    &\Longleftrightarrow \lambda^2 \leq \sum_{j=1}^n{(m_{ij}x_j)^2} \\
    &\Longrightarrow \lambda^2 \leq \sum_{j=1}^n{(m_{ij})^2} \\
    &\Longrightarrow \lambda^2 \leq \frac{1}{n}\sum_{i,j=1}^n{m_{ij}^2}
\end{align*}

Using previous results we get:

\begin{equation*}
     \sum_{i=1}^{|V_2|}{\left( \lambda_i(L_2) - \lambda_i(\overline{L_1}) \right)^2}  
     \leq |V_2|\frac{1}{|V_2|}\sum_{i,j=1}^{|V_2|}{{L_{P^*}}_{ij}^2} 
     = \lVert L_{P^*} \rVert_F^2,
\end{equation*}

with $\| X \|_F = \sqrt{\sum_{i}{\sum_{j}{|X_{ij}|^2}}}$ the Frobenius norm.


\section{Proof of Proposition \ref{prop:upper_bound_perturbation}}
\label{proof:prop:upper_bound_perturbation}

We remind that the Forbenius norm is unitarily invariant thanks to the cyclic property of the trace. 
For any $\tilde{P}\in \mathcal{O}(|V_2|)$ we have:

\begin{align*}
    \lVert L_{P^*} \rVert_F
    = & \lVert {\Pi^*}^T \overline{L_1} {\Pi^*} - L_2 \rVert_F \\
    = & \lVert {\Pi^*}^T \overline{L_1} {\Pi^*} - \tilde{P}^T L_2 \tilde{P} + \tilde{P}^T L_2 \tilde{P} - L_2 \rVert_F \\
    \leq & \lVert {\Pi^*}^T \overline{L_1} {\Pi^*} - \tilde{P}^T L_2 \tilde{P} \rVert_F + \lVert \tilde{P}^T L_2 \tilde{P} - L_2 \rVert_F.
\end{align*}

In particular if $\tilde{P}=Q_2 \overline{Q_1}^T \Pi^*$:

\begin{align*}
    \lVert \tilde{P}^T L_2 \tilde{P} - L_2 \rVert_F
    = &\lVert (Q_2 \overline{Q_1}^T \Pi^*)^T {L}_2 Q_2 \overline{Q_1}^T \Pi^* - L_2 \rVert_F \\
    = &\lVert (Q_2 \overline{Q_1}^T \Pi^*)^T Q_{2} \Lambda_2 {Q_2}^{T} Q_2 \overline{Q_1}^T \Pi^* - L_2 \rVert_F \\
    = &\lVert  {\Pi^*}^T \overline{Q_1} \Lambda_2 \overline{Q_1}^T {\Pi^*} - L_2 \rVert_F,
\end{align*}

We also have that

\begin{align*}
    \lVert {\Pi^*}^T \overline{L_1} {\Pi^*} - \tilde{P}^T {L}_2 \tilde{P} \rVert_F 
    = & \lVert Q_{1} \Lambda_1 {Q_1}^{T} - {\Pi^*} (Q_2 \overline{Q_1}^T \Pi^*)^T Q_{2} \Lambda_2 {Q_2}^{T} Q_2 \overline{Q_1}^T \Pi^* {\Pi^*}^T  \rVert_F \\
    = & \lVert \overline{\Lambda_1} - \Lambda_2 \rVert_F.
\end{align*}

Hence: $\lVert L_{P^*} \rVert_F \leq \lVert \overline{\Lambda_1} - \Lambda_2 \rVert_F + \lVert {\Pi^*}^T \overline{Q_1} \Lambda_2 \overline{Q_1}^T {\Pi^*} - L_2 \rVert_F$.

\section{Proof of Proposition \ref{prop:upper_bound}}
\label{proof:prop:upper_bound}

Denoting $\mathcal{O}(n)$ the $n$-orthogonal matrices group (orthogonal since real), we want to show that:

\begin{align*}
\label{eq:upper_bound}
    \min_{O \in \mathcal{O}(|V_2|)} \lVert \overline{L_1} - O^T L_2 O \rVert_F \leq {\|{\lambda(\overline{L_1}) - \lambda(L_2)}\|_2}
\end{align*}

We denote ${L}_i = Q_{i} \Lambda_i {Q_i}^{T}$ the eigendecomposition of the Laplacian ${L}_i$ with $\Lambda_i=\text{diag}(\lambda(L_i))$. Since $\overline{Q_1}$ is unitary and using property of Frobenius norm, we have, $\forall O \in \mathcal{O}(|V_2|)$:

\begin{align*}
    \lVert \overline{L_1} - O^T {L}_2 O \rVert_F
    & = \lVert \overline{\Lambda_1} - \overline{Q_1}^{T} O^T Q_{2} \Lambda_2 {Q_2}^{T} O \overline{Q_1} \rVert_F, \\
\end{align*}

We know that $\overline{Q_1}$ and $Q_{2}$ are orthogonal since they are respectively eigenvector matrices of symmetric matrix $\overline{L_1}$ and ${L}_2$. We therefore have:
    
    \begin{align*}
        (Q_{2}^{T} O \overline{Q_{1}})^T (Q_{2}^{T} O \overline{Q_{1}}) = \overline{Q_{1}}^{T} O^T Q_{2}Q_{2}^{T} O \overline{Q_{1}}=I_{|V_1|}
    \end{align*}

Moreover $\forall \tilde{\Pi}\in \mathcal{P}(|V_2|) \subset \mathcal{O}(|V_2|)$, if $O=Q_2 \tilde{\Pi} \overline{Q_1}^T$ then $O \in \mathcal{O}(|V_2|)$. 

Hence,

\cmmnt{
\begin{align*}
    Q_2^T O \overline{Q_1} = Q_2^T Q_2 \tilde{\Pi} \overline{Q_1}^T \overline{Q_1} = \tilde{\Pi}.
\end{align*}

and
}

\begin{align*}
    \min_{O \in \mathcal{O}(|V_2|)} \lVert \overline{L_1} - O^T L_2 O \rVert_F 
    &= \min_{O \in \mathcal{O}(|V_2|)} {\lVert \overline{\Lambda_1} - \overline{Q_1}^{T} O^T Q_{2} \Lambda_2 {Q_2}^{T} O \overline{Q_1} \rVert_F} \\
    &\leq \min_{\tilde{\Pi} \in \mathcal{P}(|V_2|)} {\lVert \overline{\Lambda_1} - \overline{Q_1}^{T} (Q_2 \tilde{\Pi} \overline{Q_1}^T)^T Q_{2} \Lambda_2 {Q_2}^{T} (Q_2 \tilde{\Pi} \overline{Q_1}^T) \overline{Q_1} \rVert_F} \\
    &= \min_{\tilde{\Pi} \in \mathcal{P}(|V_2|)} {\lVert \overline{\Lambda_1} - {\tilde{\Pi}}^T \Lambda_2 \tilde{\Pi} \rVert_F} \\
    &= \min_{\tilde{\Pi} \in \mathcal{P}(|V_2|)}{\lVert \overline{\Lambda_1} - \tilde{\Pi} \Lambda_2 \tilde{\Pi}^T \rVert_F} \\
    &= \min_{\tilde{\Pi} \in \mathcal{P}(|V_2|)}{\lVert \overline{\Lambda_1} - {\tilde{\Pi} \Lambda_2 \tilde{\Pi}^T} \rVert_F} \\
    &= \min_{\sigma \in \mathcal{S}(|V_2|)}{\| \lambda(\overline{L_1}) - \lambda(L_2)_{\sigma(1:|V_2|)} \|_2} \\
    &\leq {\|{\lambda(\overline{L_1}) - \lambda(L_2)}\|_2}
\end{align*}

with $\mathcal{S}(n)$ the permutation group of $\{1 \dots n\}$.

\section{Experimental setup for classification of graphs}
\label{app:experimental_setup}

For classification, we use standard 10-folds cross validation setup. Each dataset is divided into 10 folds such that the class proportions are preserved in each fold for all datasets. These folds are then used for cross-validation i.e, one fold serves as the testing set while the other ones compose the training set. Results are averaged over all testing sets. All figures gathered in the tables of results are build using this setup. For the dimension $d \in \llbracket 1, \max_{G \in \mathcal{D}}{|V|} \rrbracket $, representing the number of eigenvalues we keep to build the truncated GLS, we chose the percentile 95 of the distribution of graph sizes in each dataset, i.e. we truncate the 5$\%$ smallest eigenvalues. Considering weak truncation impact (see Section \ref{subsec:molecular_classification}, when we have large datasets containing large graphs, like the two REDDIT datasets, we can truncate more severely to make the problem computationally more efficient. In particular considering that GLS approached as a simple baseline more than a final graph representation for large scale usage. 

We use the support vector classifier (SVC) from scikit-learn \cite{pedregosa2011scikit}. We impose Radial Basis Function as kernel, i.e. $\mathcal{K}(\lambda(\overline{L_1}), \lambda(L_2)) = \exp{(- \gamma \lVert \lambda(\overline{L_1}) - \lambda(L_2) \lVert_2^2)}$. It is a similarity measure related to $L_2$-norm between GLS. Hence, our theoretical results remain consistent with our experiments. Hyper parameters $C$ and $\gamma$ are tuned among respectively $\{0.5, 1, 5\}$ and $\{0.0001, 0.001, 0.01, 0.1, 0.5, 1, 5\}$ for the molecular datasets, and $\{0.5, 1, 5, 25, 50\}$ and $\{0.0001, 0.001, 0.01, 0.1\}$ for the social network datasets. In practice, using a global pool for all the datasets gives equivalent results, but hyperparameter inference becomes expensive with a too large grid, in particular in a 10-fold cross validation setup. We use a nested hyperparameter search cross-validation for each of the 10 folds: in each 90\% training fold we performe a 5-fold random search cross-validation before training. We therefore avoid the problem of overfitting related to model selection that appear when using non-nested cross-validation \cite{cawley2010over}. 

\section{Characteristics of the real datasets}
\label{app:datasets_characteristics}

We use five molecular datasets and five social network datasets for the experiments \cite{KKMMN2016}. Tables \ref{tab:molecular_datasets} and \ref{tab:social_datasets} gives statistics of the differents datasets. All used datasets can be found at the following address: \url{https://ls11-www.cs.tu-dortmund.de/staff/morris/graphkerneldatasets} \cite{KKMMN2016}. 

Molecular graphs datasets are Mutag (MT), Enzymes (EZ), Proteins Full (PF), Dobson and Doig (DD) and National Cancer Institute (NCI1). In MT, the graphs are either mutagenic and not mutagenic. EZ graphs are tertiary structures of proteins from the 6 Enzyme Commission top level classes. In DD, compounds are secondary structures of proteins that are enzyme or not. PF is a subset of DD without the largest graphs. In NCI1, graphs are anti-cancer or not. The graphs of these datasets have node labels that can be leverages by graph neural networks. 

\begin{table}[h]
    \centering
    \renewcommand{\arraystretch}{1.2}
    \begin{tabular}{l|M{1.8cm} M{1.8cm} M{1.8cm} M{1.8cm} M{1.8cm}}
    
                          & MT    & EZ    & PF    & DD    & NCI1  \\
        \hline
        $\#$ graphs       & 188   & 600   & 1113  & 1178  & 4110  \\
        $\#$ classes      & 2     & 6     & 2     & 2     & 2     \\
        bias ($\%$)       & 66.5  & 16.7  & 59.6  & 58.7  & 50.0  \\
        min./max. $|V|$     & 10/28    & 2/125    & 4/620    & 30/5736   & 3/106  \\
        avg. $|V|$          & 18    & 33    & 39    & 284   & 30  \\
        avg. $|E|$          & 39    & 124   & 146   & 1431  & 65  \\
        Node attributes     & \checkmark  & \checkmark  & \checkmark  & \checkmark  & \checkmark \\
    
    \end{tabular}
    \caption{Molecular datasets statistics. Bias indicates the proportion of the dominant class.}
    \label{tab:molecular_datasets}
\end{table}

Social networks datasets are IMDB-Binary (IMBD-B), IMDB-Multi (IMDB-M), REDDIT-Binary (REDDIT-B), REDDIT-5K-Multi (REDDIT-M) and COLLAB. REDDIT-B and REDDIT-M contain graphs representing discussion thread, with edges between users (nodes) when one responded to the other's comment. Classes are the subreddit topics from which thread have originated. IMDB-B and IMDB-M contain networks of actors that appeared together within the same movie. IMDB-B contains two classes for \textit{action} or \textit{romance} genres and IMDB-M three classes for \textit{comedy}, \textit{romance} and \textit{sci-fi}. COLLAB graphs represent scientific collaborations, with edge between two researchers meaning that they co-authored a paper. Labels of the graphs correspond to subfields of Physics. The graphs of these datasets have no node attributes and therefore enable fair comparison with deep learning methods.

\begin{table}[h]
    \centering
    \renewcommand{\arraystretch}{1.2}
    \begin{tabular}{l|M{1.8cm} M{1.8cm} M{1.8cm} M{1.8cm} M{1.8cm}}
    
                          & IMDB-B    & IMDB-M   & REDDIT-B    & REDDIT-M    & COLLAB  \\
        \hline
        $\#$ graphs       & 1000   & 1500   & 2000   & 4999  & 5000 \\
        $\#$ classes      & 2     & 3     & 2     & 5     & 3       \\
        bias ($\%$)       & 50.0  & 33.3  & 50.0  & 20.0   & 52.0  \\
        min./max. $|V|$     & 12/136    & 7/89    &  3/3760  & 22/3606   & 32/492  \\
        avg. $|V|$          & 20  & 13  & 426  & 501  & 75  \\
        avg. $|E|$          & 97    & 66    & 496   & 590  & 2458  \\
        Node attributes     & \xmark & \xmark & \xmark & \xmark & \xmark \\
    
    \end{tabular}
    \caption{Social network datasets statistics. Bias indicates the proportion of the dominant class.}
    \label{tab:social_datasets}
\end{table}

\section{Additional insight on the acceptability of using truncated GLS}
\label{app:truncation}

\begin{figure}
    \centering
    \includegraphics[width=0.48\textwidth]{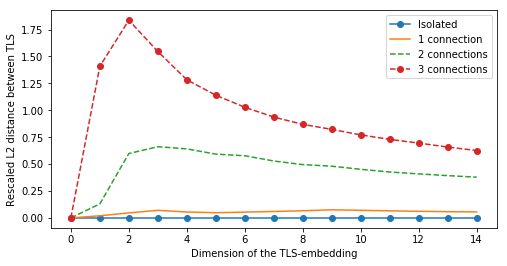}
    \includegraphics[width=0.47\textwidth]{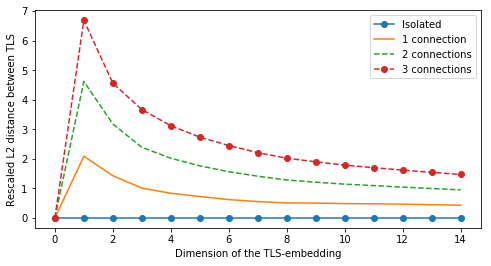}
    \caption{Illustration of the relative importance of the dimensionality of GLS-embedding, after the iterative addition of 20 new nodes with respectively 0, 1, 2 and 3 random connections with graph, for respectively synthetic a 80-nodes Erdos-Reyni graph (left) and a 28-nodes molecular graph from MUTAG dataset (right). We see that the first largest eigenvalues of the Laplacian are the most important to discriminate a graph and its perturbed version.}
    \label{fig:node_perturbations_2}
\end{figure}

Figure \ref{fig:node_perturbations_2} illustrates the reasonability of using only the highest eigenvalues of the Laplacian spectrum as whole-graph feature representation. We take the original and final graphs of the deformation-consistency test presented in Figure \ref{fig:node_perturbations}. We compute the $L_2$ distance between t-GLS with dimension $d$ and divide it by $d$, for $d$ varying from $1$ to $15$. The objective is to confirm that first eigenvalues are relatively more important to discriminate to structurally different graphs, which is the case. We note that for the Erdos-Reyni case with few connected additional nodes, first eigenvalues are not as relatively important as for the other example. In fact, adding nodes with stochastic connections is the construction process of Erdos-Reyni graphs. Hence, discriminating augmented graph from the original one is difficult based only on the structural information.

\end{document}